\documentclass[conference]{IEEEtran}
\IEEEoverridecommandlockouts
\usepackage{amsmath,amsfonts}
\usepackage{algorithmic}
\usepackage{algorithm}
\usepackage{amssymb}
\usepackage{booktabs}
\usepackage{mathrsfs}
\usepackage{float} 
\usepackage{url}
\usepackage{array}
\usepackage[caption=false,font=normalsize,labelfont=sf,textfont=sf]{subfig}
\usepackage{textcomp}
\usepackage{stfloats}
\usepackage{multirow}
\usepackage{color}
\usepackage{xcolor}
\usepackage{url}
\usepackage{verbatim}
\usepackage{graphicx}
\usepackage{cite}
\usepackage{epstopdf}
\usepackage{txfonts}

\newtheorem{theorem}{Theorem}

\newtheorem{remark}{Remark}

\newcommand{\argmin}{\mathop{{\rm arg}\min}}

\def\BibTeX{{\rm B\kern-.05em{\sc i\kern-.025em b}\kern-.08em
    T\kern-.1667em\lower.7ex\hbox{E}\kern-.125emX}}
\begin{document}

\title{Generalized Sparse Additive Model with Unknown Link Function}

\author{
\IEEEauthorblockN{Peipei Yuan}
\IEEEauthorblockA{\textit{School of Electronic Information and Communications} \\
\textit{Huazhong University of Science and Technology}\\
Wuhan, China \\
pennyyuan@hust.edu.cn}
\and
\IEEEauthorblockN{Xinge You}
\IEEEauthorblockA{\textit{School of Electronic Information and Communications} \\
\textit{Huazhong University of Science and Technology}\\
Wuhan, China\\
youxg@mail.hust.edu.cn}
\and
\IEEEauthorblockN{Hong Chen}
\IEEEauthorblockA{\textit{College of Informatics} \\
\textit{Huazhong Agricultural University}\\
Wuhan, China \\
chenh@mail.hzau.edu.cn}
\and
\IEEEauthorblockN{Xuelin Zhang}
\IEEEauthorblockA{\textit{College of Informatics} \\
	\textit{Huazhong Agricultural University}\\
	Wuhan, China \\
	zhangxuelin@webmail.hzau.edu.cn}
 \and
\IEEEauthorblockN{Qinmu Peng}
\IEEEauthorblockA{\textit{School of Electronic Information and Communications} \\
\textit{Huazhong University of Science and Technology}\\
Wuhan, China \\
pengqinmu@hust.edu.cn}
}

\maketitle

\begin{abstract}
	Generalized additive models (GAM) have been successfully applied to  high dimensional data analysis. However, most existing methods cannot simultaneously estimate the link function, the component functions and the variable interaction. To alleviate this problem, we propose a new sparse additive model, named generalized sparse additive model with unknown link function (GSAMUL), in which the component functions are estimated by B-spline basis and the unknown link function is estimated by a multi-layer perceptron (MLP) network. Furthermore, $\ell_{2,1}$-norm regularizer is used for variable selection. The proposed GSAMUL can realize both variable selection and hidden interaction. We integrate this estimation into a bilevel optimization problem, where the data is split into training set and validation set. In theory, we provide the guarantees about the convergence of the approximate procedure. In applications, experimental evaluations on both synthetic and real world data sets consistently validate the effectiveness of the proposed approach.
\end{abstract}


\section{Introduction}

Additive models and generalized additive models (GAMs) have been widely used in data analysis when exploring the nonlinear effects of the variables on the response \cite{Hastie90, Meier09, Yuan16M, Zhang2018Partially, Siems23,Chang22}. Especially for high-dimensional data, they are useful to address “the curse of dimensionality” \cite{stone1985, kohler2016}. In applications, when GAM is used to fit a dataset, it is usually assumed that the link function is a commonly used function. The link function is assumed to be the logit function for binary classification (e.g., SAM \cite{Zhao2012Sparse}, GroupSAM\cite{Chen2017Group}, CSAM\cite{yuan2023sparse}) and the identity function for linear regression (e.g., SpAM\cite{Ravikumar09}, SpMAM \cite{Chen20}). However, in some real data analysis, those commonly used link functions may be inappropriate, and incorrect link functions may lead to biased estimation of the component functions, which will affect the prediction of the model.


In machine learning literature, many additive models have been proposed for estimating the component function, which lie in the  reproducing kernel Hilbert space (RKHS) \cite{Chen2017Group,Raskutti12, Christmann16}, the space spanned by the orthogonal basis   \cite{Zhao2012Sparse,wang2011estimation, Meier09, Yin12 } or the neural networks \cite{Agarwal21, Chang22, dubey2022scalable,enouen2022sparse,radenovic2022neural,xu2023sparse}. To enhance the interpretability of prediction results, sparsity penalties are  commonly used for variable selection, e.g., $\ell_1$-norm regularizer ,  $\ell_{2,1}$-norm regularizer \cite{Ravikumar09, huang2010variable, Yin12,  Chen2017Group,Haris22}. Estimating the link function proposed earlier are mainly for generalized linear type models \cite{Weisberg1994Adapting,Carroll1998Generalized,zhang2010simultaneous,huang2014model, Zhang2015Estimation,Lin2018Efficient, Lin2022Efficient}. Carroll et al.  \cite{Carroll1998Generalized} studied the generalized partially linear single-index model with the form $\mathbb{E}(Y|X,Z)=g(f(X^T\alpha)+Z^T\beta)$, in which local linear kernel is used to estimate the parameters and the unknown function $f(\cdot)$. However, the link function $g(\cdot)$ is known. In \cite{Zhang2015Estimation}, a generalized variable coefficient model based on unknown link function $\mathbb{E}(Y|X,Z)= g(X^T\alpha(Z))$ is studied, which assumed that the inner function is linear and the coefficients are a function of the variable $Z$, and the distribution of $Y$ is an exponential family distribution.

Horowitz and Mammen \cite{Horowitz2007Rate} extended the additive models and proposed a more general additive model: $Y = g(f_1(X_1)+\cdots+f_p(X_p))+\varepsilon$, where $g(\cdot)$, $f_j(\cdot), j =1,\cdots,p$ are unknown, and $\varepsilon$ is random noise and satisfies $\mathbb{E}(\varepsilon|X)=0$. They developed an estimation process based on spline smoothing and established its optimal convergence rate. However, this model is not suitable for cases where the response variable is categorical or discrete. Actually, categorical or discrete response may also follow their respective distributions \cite{Lin2018Efficient}. For example, a researcher may be interested in predicting one of three possible discrete values. Thus, the response variable can only have three discrete values, and follow a multinomial distribution. The generalized type models can provide a distributional framework according to the type of the response. Lin et al. \cite{Lin2018Efficient} proposed a generalized additive model based on unknown link function (GAMUL):  $\mu = \mathbb{E}(Y|X)= g(f_1(X_1)+\cdots+f_p(X_p))$, $\rm{Var(Y|X)} = V(\mu)$, where $f_j(\cdot),~j=1,...,p$ are unknown component functions, $g(\cdot)$ is the unknown link function, and $V(\cdot)$ is a known variance function and determined by the type of response. For example, $V(\mu) = \mu (1-\mu)$ for binary classification, and a constant variance for regression. They proposed a quasi-likelihood backfitting method to estimate the component function and the link function. Moreover, they extended this work and proposed a generalized varying-coefficient model with unknown link and variance functions (GVULV) for large-scale data \cite{Lin2022Efficient}. However, both GAMUL and GVULV cannot simultaneously estimate the component functions and select the significant variables. Zhang and Lian\cite{Zhang2018Partially} considered the partially linear additive models with unknown link function (PLAMUL), and used polynomial spline method to estimate the link function and the component functions. This work considered the penalized estimator for variable selection and established the oracle properties of the estimators.

In recent years, the collected data has high dimension and complex structure. The above methods do not consider the variable interaction, thus usually cannot achieve satisfactory results for function estimation and variable selection. Moreover, neural networks, especially deep learning, have powerful learning capabilities and can handle high-dimensional, unstructured, and complex data. Following the research line of \cite{Zhang2018Partially}, we propose a  generalized sparse additive model with with unknown link function (GSAMUL), in which the component functions $f_j(\cdot), j = 1,\cdots,p$ are estimated via a B-spline method with $\ell_{2,1}$-norm regularizer and the link function $g(\cdot)$ is estimated via a multilayer perceptron (MLP) network.


\begin{figure}[!t]
	\centering
	\includegraphics[width= 6 cm]{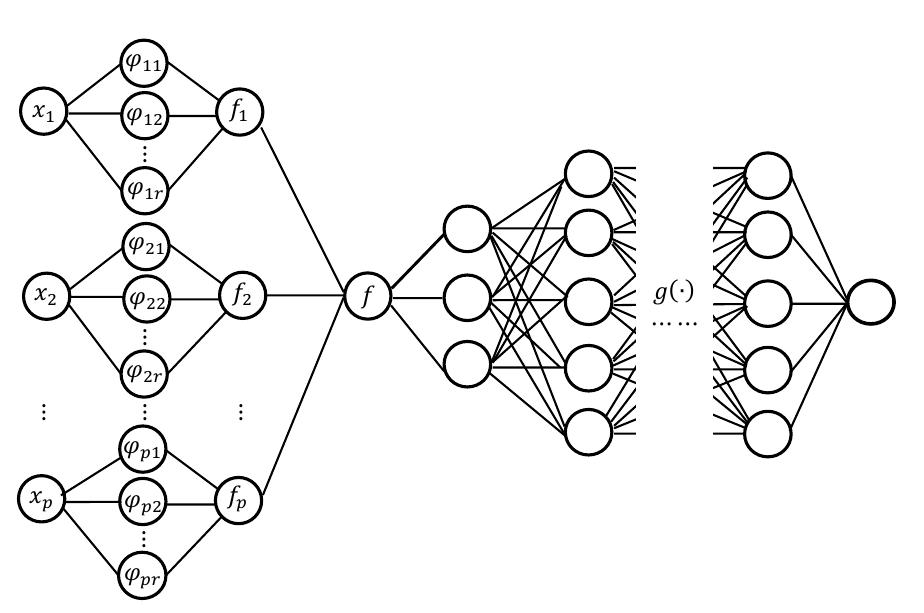}
	\caption{Overview of the proposed GSAMUL.}
	\label{Model}
\end{figure}

The contributions of this work are summarized as follows:
\begin{itemize}	
	\item Model: We proposed a new GAM with unknown link function, called GSAMUL, which is a model between a white box and a black box, and can conduct estimation and variable selection simultaneously. The additive part can learn the impact of a feature on the prediction, and the unknown link function can implicitly learn the interaction between variables. 
	\item Optimization: A bilevel optimization scheme is presented to estimate the unknown link function and the component functions in the additive part. In theory, we provide the convergence analysis of the  optimization algorithm.	
	\item Effectiveness: Empirical effectiveness of   GSAMUL is supported by experimental evaluations on simulated data and real world data. Experimental results demonstrate that GSAMUL can select the informative variables and estimate the component and link functions efficiently even if the datasets are contaminated by irrelative variables.
\end{itemize}



\section{Generalized sparse additive model} \label{method} 

\subsection{Problem Formulation}	

Let $\mathcal{Z}:=(\mathcal{X},\mathcal{Y})\subset\mathbb{R}^{p+1}$, where $\mathcal{X} \subset\mathbb{R}^{p}$ is a input space and $\mathcal{Y}\subset \mathbb{R}$ is the output space. The model studied here is
\begin{equation}\label{model}
	Y = g\Big(\sum^{p}_{j=1}f_{j}(X_{j})\Big) +\epsilon,
\end{equation}
where $X=(X_1,\cdots,X_p)^T \in \mathcal{X}$ , $Y  \in \mathcal{Y}$, $f_{j}(\cdot), j=1,\cdots,p$ is the component function, $g(\cdot)$ is the unknown link function, and $\epsilon$ is a random noise with mean zero. Here, both $g(\cdot)$ and $f_{j}(\cdot)$ are unknown quantities and needed to be estimated.

\subsection{Generalized sparse additive model with unknown link function}

This section builds an estimation procedure for the unknown functions $g(\cdot)$ and $f_j(\cdot), j = 1,\cdots,p$ in GSAMUL. 

Firstly, the component function $f_{j}(\cdot)$ is estimated via the basis-spline methods approximate. The key concept is that the unknown component functions can be expressed as a linear combination of proper basis functions.  

Denote the bounded and orthonormal basis functions on $\mathcal{X}_j$ as $\{\psi_{jk}:k=1,\cdots,\infty\}$. Then, the component functions can be written as
$	f_j(X_{j})=\sum^{\infty}_{k=1}\alpha_{jk}\psi_{jk}(X_{j})$
with coefficient $\alpha_{jk}, j=1,\cdots,p$. Actually, these basis functions is often truncated to finite dimension $d$. Then, we obtain
\begin{equation*}
	f_j(X_{j})=\sum^{d}_{k=1}\alpha_{jk}\psi_{jk}(X_{j}).
\end{equation*}

Given $n$ empirical observations  $ \mathbf{z}: = \{(x_i,y_i)\}^{n}_{i=1}\subset \mathcal{X} \times \mathcal{Y} $ with $x_i = (x_{i1},\cdots,x_{ip})^T\in \mathbb{R}^{p}$, $y_i\in \mathbb{R} $.
Let $\Psi_i=(\psi_{11}(x_{i1}),\cdots,\psi_{1d}(x_{i1}),\cdots,\psi_{p1}(x_{ip}),\cdots,\psi_{pd}(x_{ip}))\in \mathbb{R} ^{pd}$ and $\alpha = (\alpha_{11},\cdots,\alpha_{1d},\cdots,\alpha_{p1},\cdots,\alpha_{pd})^{T}\in\mathbb{R}^{pd}$. Then, the additive part $f(x_i)$ of GSAMUL can be represented as 
\begin{equation*}
	f(x_i)=\Psi_i^T\alpha.
\end{equation*}
Since neural network can accurately realize the nonlinear mapping and has large-scale computing power, we employ a MLP network to estimate the unknown link function $g(\cdot)$ in GSAMUL. Let $\theta$ be the hyper-parameter of the MLP network, and $g(\Psi_i^T\alpha;\theta)$ be the estimated model.
Then, the objective function of GSAMUL can be formulated as
\begin{equation*}
	\min\limits_{ \theta,\alpha}\frac{1}{n}\sum^{n}_{i=1}\ell(y_i,g(\Psi_i^T\alpha;\theta)),
\end{equation*}
where $\ell(\cdot)$ is the loss function. The loss function can be the least squares loss, the C-loss , cross entropy loss and so on.

In practice, it is important to determine the truly informative variables which have non-zero effects on the response. If the $j$-th variable is not truly informative, we expect that $\alpha_{j}=(\alpha_{j1},\ldots, \alpha_{jd})^T\in \mathbb{R}^d$ satisfies $\| \alpha_{j}\|_2=(\sum^d_{k=1}|\alpha_{jk}|^2) ^{\frac{1}{2}}=0$. Inspired by this, we employ the $\ell_{2,1}$-norm regularizer
\begin{equation}\label{Omega}
	\Omega(f) = \inf\Big\{ \sum^{p}_{j=1}\|\alpha_{j}\|_{2}:   f=
	\sum^{p}_{j=1}\Psi_{j}^T\alpha_{j},\alpha_{j}\in \mathbb{R}^d\Big\}
\end{equation}
as the penalty to address the sparsity of the component functions. Such coefficient-based penalties have been widely used in  sparse additive machines  \cite{Chen2017Group,Wang2021Sparse,Zhao2012Sparse}.


Apparently, model \eqref{model} is unchanged if $f_{j}$ is replaced by $f_{j}+a_j$ and $g(v)$ is replaced by $g(v+\sum_ja_j)$  for some constants $a_j, j=1,\cdots,p$ or $f_{j}$ is replaced by $cf_{j}$ and $g(v)$ is replaced by $g(v/c)$ for some constant $c\neq0$. Hence, location, sign and scale normalizations are needed to make model \eqref{model} identifable \cite{Zhang2018Partially}. Accordingly, we assume that $\alpha_{1}>0$ and $\|\alpha\|_2=1$.

Finally, the minimization problem can be described as 
\begin{equation}\label{Objective}
	\argmin \limits_{\|\alpha\|_2=1,\theta} \frac{1}{n}\sum^{n}_{i=1}\ell(y_i,g(\Psi_i^T\alpha;\theta))+\lambda \sum^{p}_{j=1}\|\alpha_{j}\|_2 , 
\end{equation}
where $\lambda>0$ is the regularization parameter.

\begin{table*}[!t]
	\scriptsize \centering \caption{Properties of models ($\surd$-the corresponding algorithm has such a property; $\times$-the corresponding algorithm hasn't such a property).}\label{RelaWorks}	\renewcommand \arraystretch{1.3}
	\begin{tabular}{p{3.2cm}p{1.8cm}p{1.8cm}p{1.8cm}p{1.8cm}p{1.8cm}p{2.6cm}} \hline
		Methods		&SpAM\cite{Ravikumar09}   &GLMUL\cite{Muggeo2008Fitting}					&GAMUL\cite{Lin2018Efficient}	&PLAMUL\cite{Zhang2018Partially} 				&GSAMUL(Ours)		\tabularnewline\hline		
		Component function &B-spline&Linear&Unknown&Unknown&Unknown    \tabularnewline
		Link function &Identify&Unknown&Unknown&Unknown&Unknown    \tabularnewline
		Regularizer	&$\ell_{1}-$norm&$-$&$-$&$\ell_{1}-$norm&$\ell_{2,1}-$norm   \tabularnewline	
		Optimization Framework &Single-level
		&Single-level
		&Single-level
		&Single-level
		&Bi-level \tabularnewline	
		Sparseness &$\surd$&$\times$&$\times$&$\surd$&$\surd$   \tabularnewline	
		Variable interaction &$\times$&$\times$&$\times$&$\times$&$\surd$  \tabularnewline \hline
	\end{tabular}
\end{table*}

Given the training set $Z^{tra}=\{(x_i,y_i)\}^{n}_{i=1}$ and validation set $Z^{val}=\{(x_i,y_i)\}^{m}_{i=1}$, our GSAMUL can be formulated as the following bilevel optimization scheme:

\textbf{Outer Problem (based on validation set $Z^{val}$):} The optimal parameter $\hat{\theta}$ is calculated by minimizing the objective:
\begin{equation}\label{Outer}
	\hat{\theta}  = \argmin \limits_{\theta} \frac{1}{m}\sum^{m}_{i=1}\ell(y_i,g(\Psi_i^T\hat{\alpha}  ;\theta)),
\end{equation}
where $\hat{\alpha}$ is the minimization of the following problems:

\textbf{Inner Problem (based on training set $Z^{tra}$):}
\begin{equation}\label{Inner}
	\hat{\alpha} (\theta) = \argmin \limits_{\|\alpha\|=1} \frac{1}{n}\sum^{n}_{i=1}\ell(y_i,g(\Psi_i^T\alpha;\hat{\theta} ))+\lambda \sum^{p}_{j=1}\|\alpha_{j}\|_2 .
\end{equation}

\begin{remark}
	Compared with the related methods. The GSAMUL is an extention of SpAM from known link function to unknown link function. If the link function is identify function, GSAMUL reduces to SpAM when using the least squares loss. Compared with GLMUL and GAMUL, GSAMUL can employ  
	a penalty term to the objective function and conduct both functin estimation and variable selection. In comparison with PLAMUL, a MLP network is employed to estimate the unknown link function, whcih can implicitly learn the interaction between variables. Furthermore, the component functions and unkonwn link function are estimated  by means of a bilevel optimization problem, where the data is split into the training set and the validation set. Now we state Table \ref{RelaWorks} to briefly compare GSAMUL with the related works of GAMs including SpAM\cite{Ravikumar09}, GLMUL\cite{Muggeo2008Fitting}, GAMUL\cite{Lin2018Efficient}, PLAMUL\cite{Zhang2018Partially}, we can find the proposed approach is new and enjoys nice properties, i.e., sparseness and variable interaction.
\end{remark}

\begin{remark}
   The bilevel optimization problem in \eqref{Outer} and \eqref{Inner} can be solved via an online strategy in \cite{shu2023cmw,shu2019meta}. Furthermore, the informative variable set is obtained as
		$\mathcal{J}_{\mathbf{z}}=\{j:\|\hat{\alpha}_{j}\|_2\geq v_n \}$,
	where $v_n$ is a positive threshold obtained via a stability-based selection strategy \cite{Sun2013}. The detail optimization algorithm is provided in Section \ref{Optimization}.
\end{remark}
\begin{remark}
	The component functions in additive part can also be estimated via the neural network, which can be considered as a generalization of neural additive models (GNAMs). And, the optimization process and theoretical analysis can be implemented in the similar way.    
\end{remark}


\section{Optimization algorithm for GSAMUL}\label{Optimization}

The general optimization of GSAMUL is stated in the following two steps.


\subsection{Step 1: Estimating $\hat{\alpha}$ and $\hat{\theta}$ via bilevel optimization scheme}

Calculating the optimal $\hat{\alpha}$ and $\hat{\theta}$ requires two nested loops of optimization. To guarantee the efficiency of GSAMUL, an online strategy in \cite{shu2023cmw,shu2019meta} is  employed to update the parameters $\theta$ and $\alpha$ through a single optimization loop, respectively.

As general network training tricks, we employ stochastic gradient descent(SGD) to optimize the additive part in \eqref{Inner}. 
Specifically, in each iteration of training, we random sample a mini-batch of training samples $\{(x_i,y_i)\}^{n'}_{i=1}$, where $n'$ is the sample number in the mini-batch. Then,  updating $\alpha_j, j=1,\cdots,p$ can be formulated by moving the current $\alpha^{(t)}$ along the descent direction of the objective in \eqref{Inner} on a mini-batch training data: 
\begin{equation}\label{alpha_hat}
	\bar{\alpha}	^{(t+1)}_j(\theta) =  (1-    \frac{\eta\lambda }{\|\alpha^{(t)}_{j}\|_2}   )\alpha_j^{(t)} -  \frac{\eta}{n'} 	\sum^{n'}_{i=1}\nabla_{\alpha_j}\ell(y_i,g(\Psi_i^T\alpha;\theta))\Big|_{\alpha_j^{(t)}},
\end{equation}
where  $j = 1,\cdots,p$ and $\eta$ is the step size.

After receiving the feedback of the parameter updating $\bar{\alpha}^{(t+1)}(\theta)$ form \eqref{alpha_hat}, the parameter $\theta$ of the MLP network can be updated by moving the current parameter $\theta^{(t)}$ along the objective gradient of  \eqref{Outer} calculated on the validation set $Z^{val}$:
\begin{equation}\label{theta_t}
	\theta^{(t+1)} = \theta^{(t)}  -   \frac{\nu}{m} 	\sum^{m}_{i=1}\nabla_{\theta}\ell(y_i,g(\Psi_i^T\bar{\alpha}^{(t+1)};\theta))\Big|_{\theta^{(t)}},
\end{equation}
where $\nu$ is the step size.

Then,  $\theta^{(t+1)}$ is employed to update the parameter $\alpha_j$:

	\begin{equation}\label{alpha_t}
		\alpha^{(t+1)}_j(\theta) =  (1- \eta\frac{\lambda }{\|\alpha^{(t)}_{j}\|_2} )\alpha_j^{(t)} -  \frac{\eta}{n} 	\sum^{n}_{i=1}\nabla_{\alpha}\ell(y_i,g(\Psi_i^T\alpha;\theta^{(t+1)} ))  \Big|_{\alpha_j^{(t)}},
	\end{equation}

\subsection{Step 2: Selecting active variables $\mathcal{J}_{\mathbf{z}}$}
Here, we adopt a stability-based selection strategy \cite{Sun2013} to obtain a stable result for variable selection and estimation.

First, randomly divide the training set into two subsets. Given a constant $v_n$, two variable sets $J_{k1}$ and $J_{k2}$ are produced for each subset with $k$-th splitting.

Then, derive the threshold $v_n$ via maximizing the  quantity \begin{equation}\label{Cohen}
		\frac{1}{T} \sum_{k=1}^T  \kappa(J_{k1},J_{k2}),
\end{equation} 
where $T$ is the number of partition in the data-adaptive scheme, and $\kappa(\cdot,\cdot)$ is the Cohens kappa coefficient between two informative sets \cite{Sun2013}.  

Finally, obtain the active variables: $\mathcal{J}_{\mathbf{z}}=\{j:\|\hat{\alpha}^{(j)}\|_2\geq v_n \}$.

The optimization algorithm of the proposed GSAMUL is summarized in Algorithm \ref{algorithm1}. 

\begin{algorithm}[H]
	\caption{: Optimization Algorithm of GSAMUL.}  \label{algorithm1}
	\begin{algorithmic}
		\STATE\textbf{Input:}
		Training set $Z^{tra}$, validation set $Z^{val}$, batch size $n'$, Max iterations $T$, $\lambda$, $\eta$ and $\nu$;
		\STATE\textbf{Initialization:} Initialize $\alpha^{(0)}$ and $\theta^{(0)}$ via the uniform distribution $U(0,1)$ and $t=0;$
		\FOR {$t=0$ to $T-1$ } 
		\STATE 1. $\{x,y\}\leftarrow $ SampleMiniBatch $(\mathcal{Z}, n')$
		\STATE 2. Calculate  $\bar{\alpha}_j^{(t)}$ of the additive part via  \eqref{alpha_hat};
		\STATE 3. Update $\theta^{(t+1)}$ via \eqref{theta_t};
		\STATE 4. Update $\alpha^{(t+1)}_{j}$ via  \eqref{alpha_t};
		\ENDFOR		
		\STATE $\hat{\alpha}_{j} = \alpha^{(T)}_{j}, ~j = 1,...,p$, $\hat{\theta} = \theta^{(T)}$
		\STATE\textbf{Variable Selection:} $\mathcal{J}_{\mathbf{z}}=\{j:\|\hat{\alpha}_{j}\|_2 \geq v_n \}$, where $v_n$ is obtained by maximizing \eqref{Cohen};
		\STATE\textbf{Output:} $\hat{\alpha}$,  $\hat{\theta}$  and $\mathcal{J}_{\mathbf{z}}$ 
	\end{algorithmic}
\end{algorithm}

\section{Convergence analysis}\label{Convergence}

Since GSAMUL involves the optimization of bilevel objectives, we prove that GSAMUL converges to a critical points under some mild conditions in Theorem 1 and 2.


Suppose that we have a validation dataset with $m$ samples $Z^{val}=\{(x_i,y_i)\}^{m}_{i=1}$, and the overall validation loss is
\begin{equation*} 
	\mathcal{L}^{val}(\hat{\alpha},\theta) = \frac{1}{m}\sum^{m}_{i=1}L^{val}_i(\hat{\alpha},\theta),
\end{equation*}
where 
$L^{val}_i(\hat{\alpha},\theta)=\ell(y_i, g(\Psi_i^T\hat{\alpha}, \theta))$, $\hat{\alpha}$ is the parameter of the additive part, and $\theta$ is the parameter of the MLP network. Suppose that we have a training dataset with $n$ samples $Z^{tra}=\{(x_i,y_i)\}^{n}_{i=1}$, and  $\hat{\alpha}$ is learned via the minimization of the following structural risk:
\begin{equation*} 
	\mathcal{L}^{tr}(\alpha,\hat{\theta}) = \frac{1}{n}\sum^{n}_{i=1}L^{tra}_i(\alpha,\hat{\theta}) + \lambda \sum^{p}_{j=1}\|\alpha_j\|_2,
\end{equation*}
where $L^{tr}_i(\alpha,\hat{\theta})=\ell(y_i, g(\Psi_i^T\alpha, \hat{\theta}))$, $\alpha =(\alpha^T_1,...,\alpha^T_p)^T$, $\lambda$ is the regularization parameter. 

\begin{theorem}\label{Theorem1}
	Suppose the loss function $\ell$ is Lipschitz smooth with constant $L$ and has $\rho$-bounded gradient with respect to training and validation data. Let the learning rate $\eta_t, \nu_t, 1\leq t\leq T$  be monotonically descent sequences, and satisfy $\eta_t = \min\{\frac{1}{L},\frac{c}{\sqrt{T}}\}$, $\nu_t=\min\{\frac{1}{L},\frac{c}{\sqrt{T}}\}$ for some $c>0$ such that $\frac{\sqrt{T}}{c}>L$. Furthermore, let $\eta_t, \nu_t$ satisfy that $\sum^{\infty}_{t=1}\eta_t = \infty$, $\sum^{\infty}_{t=1}{\eta_t}^2 < \infty$, $\sum^{\infty}_{t=1}\nu_t = \infty$, and $\sum^{\infty}_{t=1}{\nu_t}^2 < \infty$. Then, the MLP net can achieve $\mathbb{E}[\|\nabla_{\theta}\mathcal{L}^{val}(\bar{\alpha}^{(t)};\theta^{(t)})\|^2_2] \leq \epsilon$ in $\mathcal{O}(1/\epsilon^2)$ steps. More specifically, 
	\begin{equation*} 
		\mathbb{E}[\|\nabla_{\theta}\mathcal{L}^{val}(\bar{\alpha}^{(t)};\theta^{(t)})\|^2_2] \leq  \mathcal{O}(\frac{C}{\sqrt{T}}),
	\end{equation*}
	where $C$ is a constant independent of the iteration process.
\end{theorem}

\noindent{\bf{Proof}.}  
The updating equation of $\theta$ in each iteration is 
\begin{equation} \label{theta_j}
	\theta^{(t+1)} = \theta^{(t)}  -   \frac{\nu}{m} 	\sum^{m}_{i=1}\nabla_{\theta} 
	L^{val}_i(\bar{\alpha}^{(t+1)},\theta)\Big| _{\theta^{(t)}}.
\end{equation}
Under the validation dataset $\Xi_t$ \eqref{theta_j} can be rewritten as:
\begin{equation*} 
	\theta^{(t+1)} = \theta^{(t)}- \nu_t	\nabla_{\theta} 
	\mathcal{L}^{val}(\bar{\alpha}^{(t+1)},\theta)\Big|_{\Xi_t}.
\end{equation*}

Since the validation dataset $\Xi_t$ is drawn uniformly from the entire dataset, the above update equation can be written as:
\begin{equation*} 
	\theta^{(t+1)} = \theta^{(t)}- \nu_t[	\nabla_{\theta} 
	\mathcal{L}^{val}(\bar{\alpha}^{(t+1)},\theta^{(t)}) +\xi^{(t)}],
\end{equation*}
where $\xi^{(t)}= \nabla_{\theta} 
\mathcal{L}^{val}(\bar{\alpha}^{(t+1)},\theta)| _{\Xi_t}-\nabla_{\theta} 
\mathcal{L}^{val}(\bar{\alpha}^{(t+1)},\theta)$. Since $\Xi_t$ are drawn independent and identically distributed (i.i.d)	with a finite number of samples, then $\xi^{(t)}$  is a random variable with finite variance. Hence, we have $\mathbb{E}[\xi^{(t)}] =0$,  $\mathbb{E}[\|\xi^{(t)}\|^2_2]<\sigma^2$.

Observe that
\begin{align} \label{Theorem_1}
	&~~~\mathcal{L}^{val}(\bar{\alpha}^{(t+1)};\theta^{(t+1)}) - \mathcal{L}^{val}(\bar{\alpha}^{(t)};\theta^{(t)})~\nonumber  \\
	&=\Big\{\mathcal{L}^{val}(\bar{\alpha}^{(t+1)};\theta^{(t+1)}) - \mathcal{L}^{val}(\bar{\alpha}^{(t+1)};\theta^{(t)}) \Big\}\\
 &~~~+\Big\{\mathcal{L}^{val}(\bar{\alpha}^{(t+1)};\theta^{(t)})- \mathcal{L}^{val}(\bar{\alpha}^{(t)};\theta^{(t)})\Big \}.\nonumber
\end{align} 
For the first term, based on the Lipschitz smoothness of the validating loss function $\mathcal{L}^{val}(\bar{\alpha}^{(t+1)};\theta^{(t+1)})$, we have 
	\begin{align} \label{Theorem_2}
		&~~~\mathcal{L}^{val}(\bar{\alpha}^{(t+1)};\theta^{(t+1)}) - \mathcal{L}^{val}(\bar{\alpha}^{(t+1)};\theta^{(t)}) \nonumber \\
		&\leq \Big\langle \nabla_{\theta}\mathcal{L}^{val}(\bar{\alpha}^{(t+1)};\theta^{(t)}), \theta^{(t+1)}-\theta^{(t)} \Big\rangle + \frac{L}{2}\Big\|\theta^{(t+1)} -\theta^{(t)} \Big\|^2_2 \nonumber \\
		&~~~+ \frac{L\nu_t^2}{2}\Big\|\nabla_{\theta} 
		\mathcal{L}^{val}(\bar{\alpha}^{(t+1)},\theta^{(t)}) +\xi^{(t)} \Big\|^2_2  \\
		&= (\frac{L\nu_t^2}{2}-\nu_t)\Big\|\nabla_{\theta} \mathcal{L}^{val}(\bar{\alpha}^{(t+1)};\theta^{(t)})\Big\|^2_2 +\frac{L\nu_t^2}{2}\Big\|\xi^{(t)}\Big\|^2_2 \nonumber  \nonumber \\
		&~~~+ (L\nu_t^2 - \nu_t) \Big\langle \nabla_{\theta} \mathcal{L}^{val}(\bar{\alpha}^{(t+1)};\theta^{(t)}) ,\xi^{(t)}\Big\rangle. & \nonumber 
	\end{align}
For the second term, we can obtain
	\begin{align} \label{Theorem_3}
		&~~~ \mathcal{L}^{val}(\bar{\alpha}^{(t+1)};\theta^{(t)}) - \mathcal{L}^{val}(\bar{\alpha}^{(t)};\theta^{(t)})\nonumber\\
		&\leq \Big\langle \nabla_{\alpha}\mathcal{L}^{val}(\bar{\alpha}^{(t)};\theta^{(t)}), \bar{\alpha}^{(t+1)}-\bar{\alpha}^{(t)} \Big\rangle + \frac{L}{2}\Big\|\bar{\alpha}^{(t+1)} -\bar{\alpha}^{(t)} \Big\|^2_2    \\
		&= \sum^{p}_{j=1}\Big\{ \Big\langle \nabla_{\alpha_j}\mathcal{L}^{val}(\bar{\alpha}^{(t)};\theta^{(t)}), \bar{\alpha}_j^{(t+1)}-\bar{\alpha}_j^{(t)} \Big \rangle + \frac{L}{2}\Big\|\bar{\alpha}_j^{(t+1)} -\bar{\alpha}_j^{(t)} \Big\|^2_2 \Big\}.&\nonumber 
	\end{align}
According the updating equation of $\bar{\alpha}_j^{(t+1)}$, we have 
\begin{equation*}
	\bar{\alpha}_j^{(t+1)} = \bar{\alpha}_j^{(t)}  -\eta_t \nabla_{\alpha_j} \mathcal{L}^{tr}(\alpha^{(t)};\theta^{(t)}) \Big|_{\Phi_t},
\end{equation*}
where $\Phi_t$ is drawn randomly and  uniformly form the training dataset in the $t$-th iteration and $\nabla_{\alpha_j} \mathcal{L}^{tr}(\alpha^{(t)};\theta^{(t)})= \nabla_{\alpha_j} \mathcal{L}_{n'}^{tr}(\alpha^{(t)};\theta^{(t)})+\lambda \frac{\alpha_j^{(t)}}{\|\alpha_j^{(t)}\|_2}$.

Then, we can rewrite the updating equation  as 
\begin{equation*}
	\bar{\alpha}_j^{(t+1)} = \bar{\alpha}_j^{(t)}  -\eta_t[\nabla_{\alpha_j} \mathcal{L}_{n'}^{tr}(\alpha^{(t)};\theta^{(t)})+\varphi^{(t)}+\lambda \frac{\alpha_j^{(t)}}{\|\alpha_j^{(t)}\|_2}],
\end{equation*}
where $\varphi^{(t)}= \nabla_{\alpha_j} \mathcal{L}_{n'}^{tr}(\alpha^{(t)};\theta^{(t)})\Big|_{\Phi_t}-\nabla_{\alpha_j} \mathcal{L}_{n'}^{tr}(\alpha^{(t)};\theta^{(t)})$. Since $\Phi_t$ is drawn i.i.d with a finte number of samples,  $\varphi^{(t)}$ are i.i.d  random variables with finte variance, and $ \mathbb{E}[\varphi^{(t)}]=0$,  $\mathbb{E}[\|\varphi^{(t)}\|^2_2]\leq \sigma^2$. 

Therefore, we have
\allowdisplaybreaks[4]

	\begin{align}\label{Theorem_4}
		&~~~\Big\langle \nabla_{\alpha_j}\mathcal{L}^{val}(\bar{\alpha}^{(t)};\theta^{(t)}), \bar{\alpha}_j^{(t+1)}-\bar{\alpha}_j^{(t)} \Big\rangle + \frac{L}{2}\Big\|\bar{\alpha}_j^{(t+1)} -\bar{\alpha}_j^{(t)} \Big\|^2_2 \nonumber\\
		&=
		\Big\langle \nabla_{\alpha_j}\mathcal{L}^{val}(\bar{\alpha}^{(t)}; \theta^{(t)}), -\eta_t\nabla_{\alpha_j} \mathcal{L}_{n'}^{tr}(\alpha^{(t)};\theta^{(t)}) \Big\rangle +\frac{L\eta^2_t}{2}\Big\| \varphi^{(t)}\Big\|^2_2  \nonumber\\
		&~~~+ \Big\langle\nabla_{\alpha_j}\mathcal{L}^{val}(\bar{\alpha}^{(t)}; \theta^{(t)}) , -\eta_t\varphi^{(t)} \Big\rangle + \Big\langle\nabla_{\alpha_j}\mathcal{L}^{val}(\bar{\alpha}^{(t)}; \theta^{(t)}) , -\eta_t\lambda \frac{\alpha_j^{(t)}}{\|\alpha_j^{(t)}\|_2} \Big\rangle   \nonumber \\
		&~~~ + \frac{L\eta^2_t}{2}\lambda^2 +\frac{L\eta^2_t}{2} \Big\|\nabla_{\alpha_j} \mathcal{L}_{n'}^{tr}(\alpha^{(t)};\theta^{(t)}) \Big\|^2_2    + L\eta^2_t \Big\langle \varphi^{(t)}, \lambda \frac{\alpha_j^{(t)}}{\|\alpha_j^{(t)}\|_2}\Big\rangle \\
		&~~~ +L\eta^2_t \Big\langle \nabla_{\alpha_j} \mathcal{L}_{n'}^{tr}(\alpha^{(t)};\theta^{(t)}) , \lambda \frac{\alpha_j^{(t)}}{\|\alpha_j^{(t)}\|_2}\Big\rangle + L\eta^2_t \Big\langle  \nabla_{\alpha_j} \mathcal{L}_{n'}^{tr}(\alpha^{(t)};\theta^{(t)}) , \varphi^{(t)}\Big\rangle  \nonumber\\
		&\leq   
  \eta_t\Big\|\nabla_{\alpha_j}\mathcal{L}^{val}(\bar{\alpha}^{(t)}; \theta^{(t)})\Big\|_2\Big\| \nabla_{\alpha_j} \mathcal{L}_{n'}^{tr}(\alpha^{(t)};\theta^{(t)})\Big\|_2  +\lambda\eta_t \Big\|\nabla_{\alpha_j}\mathcal{L}^{val}(\bar{\alpha}^{(t)}; \theta^{(t)}) \Big\|\nonumber\\
		&~~~ + \Big\langle\nabla_{\alpha_j}\mathcal{L}^{val}(\bar{\alpha}_j^{(t)}; \theta^{(t)}) , -\eta_t\varphi^{(t)} \Big\rangle +\frac{L\eta^2_t}{2}\Big\| \varphi^{(t)}\Big\|^2_2   + \frac{L\eta^2_t}{2}\lambda^2 \nonumber\\
		&~~~  +\frac{L\eta^2_t}{2} \Big\|\nabla_{\alpha_j} \mathcal{L}_{n'}^{tr}(\alpha^{(t)};\theta^{(t)}) \Big\|^2_2  +L\lambda\eta^2_t \Big\|\nabla_{\alpha_j} \mathcal{L}_{n'}^{tr}(\alpha^{(t)};\theta^{(t)}) \Big\|_2  \nonumber\\
		&~~~+L\eta^2_t \Big\langle \varphi^{(t)}, \lambda \frac{\alpha_j^{(t)}}{\|\alpha_j^{(t)}\|_2}\Big\rangle + L\eta^2_t \Big\langle  \nabla_{\alpha_j} \mathcal{L}_{n'}^{tr}(\alpha^{(t)};\theta^{(t)}) , \varphi^{(t)}\Big\rangle,& \nonumber
	\end{align}

where the last inequality holds since $\langle a,b\rangle\leq\|a\|_2\|b\|_2$.

Combining \eqref{Theorem_1}-\eqref{Theorem_4}, we have 
\begin{align*}  
	&~~~\mathcal{L}^{val}(\bar{\alpha}^{(t+1)};\theta^{(t+1)}) - \mathcal{L}^{val}(\bar{\alpha}^{(t)};\theta^{(t)})  \\
	&\leq 	  (\frac{L\nu_t^2}{2}-\nu_t)\Big\|\nabla_{\theta} \mathcal{L}^{val}(\bar{\alpha}^{(t+1)};\theta^{(t)})\Big\|^2_2 +\frac{L\nu_t^2}{2}\Big\|\xi^{(t)}\Big\|^2_2   \\
	&~~~+ (L\nu_t^2 - \nu_t) \Big\langle \nabla_{\theta} \mathcal{L}^{val}(\bar{\alpha}^{(t+1)};\theta^{(t)}) ,\xi^{(t)}\Big\rangle +\frac{L\eta^2_t}{2}\Big\| \varphi^{(t)}\Big\|^2_2 \\
	&~~~+  \sum^{p}_{j=1}\Big \{\eta_t\Big\|\nabla_{\alpha_j}\mathcal{L}^{val}(\bar{\alpha}^{(t)}; \theta^{(t)})\Big\|_2\Big\| \nabla_{\alpha_j} \mathcal{L}_{n'}^{tr}(\alpha^{(t)};\theta^{(t)})\Big\|_2 \\ 	
	&~~~+ \Big\langle\nabla_{\alpha_j}\mathcal{L}^{val}(\bar{\alpha}^{(t)}; \theta^{(t)}) , -\eta_t\varphi^{(t)} \Big\rangle  +\lambda\eta_t \Big\|\nabla_{\alpha_j}\mathcal{L}^{val}(\bar{\alpha}^{(t)}; \theta^{(t)}) \Big\| \\
	&~~~ +\frac{L\eta^2_t}{2} \Big\|\nabla_{\alpha_j} \mathcal{L}_{n'}^{tr}(\alpha^{(t)};\theta^{(t)}) \Big\|^2_2    + \frac{L\eta^2_t}{2}\lambda^2+L\eta^2_t \Big\langle \varphi^{(t)}, \lambda \frac{\alpha_j^{(t)}}{\|\alpha_j^{(t)}\|_2}\Big\rangle\\
	&~~~+L\lambda\eta^2_t \Big\|\nabla_{\alpha_j} \mathcal{L}_{n'}^{tr}(\alpha^{(t)};\theta^{(t)}) \Big\|_2 + L\eta^2_t \Big\langle  \nabla_{\alpha_j} \mathcal{L}_{n'}^{tr}(\alpha^{(t)};\theta^{(t)}) , \varphi^{(t)}\Big\rangle\Big\}  &
\end{align*}  

Furthermore, taking expectation with respect to $\xi^{(t)}$ and $\varphi^{(t)}$ on both sides of  the above inequality,  we can deduce that 

\begin{align} \label{expet_val}
	&~~~(\nu_t - \frac{L\nu_t^2}{2})\Big\|\nabla_{\theta} \mathcal{L}^{val}(\bar{\alpha}^{(t+1)};\theta^{(t)})\Big\|^2_2\nonumber \\
	&\leq  \mathcal{L}^{val}(\bar{\alpha}^{(t)};\theta^{(t)})-\mathcal{L}^{val}(\bar{\alpha}^{(t+1)};\theta^{(t+1)})+\frac{L\nu_t^2}{2}\sigma^2 \nonumber \\ 
	&~~~ +\sum^{p}_{j=1}\Big \{  \rho\eta_t\Big\|\nabla_{\alpha_j}\mathcal{L}^{val}(\bar{\alpha}^{(t)}; \theta^{(t)})\Big\|_2 +\lambda\eta_t \Big\|\nabla_{\alpha_j}\mathcal{L}^{val}(\bar{\alpha}^{(t)}; \theta^{(t)}) \Big\|\nonumber \\
	&~~~+\frac{L\eta^2_t}{2}\rho^2+\frac{L\eta^2_t}{2}\lambda^2+\frac{L\eta^2_t}{2}\sigma^2 +L\lambda\rho\eta^2_t \Big\}\\
	&= \mathcal{L}^{val}(\bar{\alpha}^{(t)};\theta^{(t)})-\mathcal{L}^{val}(\bar{\alpha}^{(t+1)};\theta^{(t+1)})+\frac{L\nu_t^2}{2}\sigma^2 \nonumber \\      
	&~~~ +  \sum^{p}_{j=1}(\rho+\lambda)\eta_t\Big\|\nabla_{\alpha_j} \mathcal{L}^{val}(\bar{\alpha}^{(t)}; \theta^{(t)})\Big\|_2 +\frac{pL\eta^2_t}{2}(\rho^2+\lambda^2+\sigma^2+2\rho\lambda) .\nonumber 
\end{align}

Summing up \eqref{expet_val} over $t=1,\cdots,T$ on both sides, we have 

\begin{align*} 
	&~~~\sum^{T}_{t=1}(\nu_t - \frac{L\nu_t^2}{2})\Big\|\nabla_{\theta} \mathcal{L}^{val}(\bar{\alpha}^{(t+1)};\theta^{(t)})\Big\|^2_2\\
	&=  \mathcal{L}^{val}(\bar{\alpha}^{(1)};\theta^{(1)})-\mathcal{L}^{val}(\bar{\alpha}^{(T+1)};\theta^{(T+1)})+ \sum^{T}_{t=1}\frac{L\nu_t^2}{2}\sigma^2 \\      
	&~~~ +  (\rho+\lambda)\sum^{T}_{t=1}\eta_t\sum^{p}_{j=1}\Big\|\nabla_{\alpha_j} \mathcal{L}^{val}(\bar{\alpha}^{(t)}; \theta^{(t)})\Big\|_2\\    
	&~~~+\frac{pL(\rho^2+\lambda^2+\sigma^2+2\rho\lambda)}{2} \sum^{T}_{t=1}\eta^2_t\\
	&\leq \mathcal{L}^{val}(\bar{\alpha}^{(1)};\theta^{(1)})+\frac{pL(\rho^2+\lambda^2+\sigma^2+2\rho\lambda)}{2} \sum^{T}_{t=1}\eta^2_t\\      
	&~~~ + \frac{L\sigma^2}{2}\sum^{T}_{t=1} \nu_t^2+  (\rho+\lambda)\sum^{T}_{t=1}\eta_t\sum^{p}_{j=1}\Big\|\nabla_{\alpha_j} \mathcal{L}^{val}(\bar{\alpha}^{(t)}; \theta^{(t)})\Big\|_2< \infty,&
\end{align*}

where the last inequality holds since $\sum^{T}_{t=1}\eta^2_t<\infty$, $\sum^{T}_{t=1}\nu^2_t<\infty$ and $\sum^{T}_{t=1}\eta_t\sum^{p}_{j=1} \|\nabla_{\alpha_j} \mathcal{L}^{val}(\bar{\alpha}^{(t)}; \theta^{(t)}) \|_2<\infty$.

Therefore, based on the above inequality, we have 

\begin{align*} 	
	&~~~	\min\limits_{t} \mathbb{E}[\|\nabla_{\theta} \mathcal{L}^{val}(\bar{\alpha}^{(t+1)};\theta^{(t)})\|^2_2]\\
	&\leq \frac{\sum^{T}_{t=1}(\nu_t - \frac{L\nu_t^2}{2})\mathbb{E}[\|\nabla_{\theta} \mathcal{L}^{val} (\bar{\alpha}^{(t+1)};\theta^{(t)})\|^2_2] }{\sum^{T}_{t=1}(\nu_t - \frac{L\nu_t^2}{2})}\\
	&\leq \frac{2\sum^{T}_{t=1}(\nu_t - \frac{L\nu_t^2}{2})\mathbb{E}[\|\nabla_{\theta} \mathcal{L}^{val}(\bar{\alpha}^{(t+1)};\theta^{(t)})\|^2_2] }{\sum^{T}_{t=1}\nu_t }\\	
	&\leq \frac{2\sum^{T}_{t=1}(\nu_t - \frac{L\nu_t^2}{2})\mathbb{E}[\|\nabla_{\theta} \mathcal{L}^{val}(\bar{\alpha}^{(t+1)};\theta^{(t)})\|^2_2] }{T}\max\{L,\dfrac{\sqrt{T}}{c}\}\\
	&= \mathcal{O}(\frac{C}{\sqrt{T}}).	&
\end{align*}

The second inequality holds since $\sum^{T}_{t=1}(2\nu_t - L\nu_t^2)=\sum^{T}_{t=1}\nu_t(2- L\nu_t)\geq\sum^{T}_{t=1}\nu_t$.  
Therefore, we can deduce that the proposed algorithm can always achieve $\min\limits_{t} \mathbb{E}[\|\nabla_{\theta} \mathcal{L}^{val}(\bar{\alpha}^{(t+1)};\theta^{(t)})\|^2_2]\leq\mathcal{O}(\frac{C}{\sqrt{T}})$ in $T$ steps. This completes the proof of Theorem \ref{Theorem1}. $\blacksquare $

\begin{theorem}\label{Theorem2}
	Suppose the loss function $\ell$ is Lipschitz smooth with constant $L$ and has $\rho$-bounded gradient with respect to training and validation data. Let the learning rate $\eta_t, \nu_t, 1\leq t\leq T$  be monotonically descent sequences, and satisfy $\eta_t = \min\{\frac{1}{L},\frac{c}{\sqrt{T}}\}$, $\nu_t=\min\{\frac{1}{L},\frac{c}{\sqrt{T}}\}$ for some $c>0$ such that $\frac{\sqrt{T}}{c}>L$. Furthermore, let $\eta_t, \nu_t$ satisfy that $\sum^{\infty}_{t=1}\eta_t = \infty$, $\sum^{\infty}_{t=1}{\eta_t}^2 < \infty$, $\sum^{\infty}_{t=1}\nu_t = \infty$, and $\sum^{\infty}_{t=1}{\nu_t}^2 < \infty$. Then, the GSAMUL can achieve $\mathbb{E}[\|\nabla_{\alpha_j}\mathcal{L}^{tr}(\alpha^{(t)};\theta^{(t)})\|^2_2] \leq \epsilon$ in $\mathcal{O}(1/\epsilon^2)$ steps. More specifically, 
	\begin{equation*} 
		\mathbb{E}[\|\nabla_{\alpha_j}\mathcal{L}^{tr}(\alpha^{(t)};\theta^{(t)})\|^2_2] \leq  \mathcal{O}(\frac{C}{\sqrt{T}}),
	\end{equation*}
	where $C$ is a constant independent of the iteration process.
\end{theorem}

\noindent{\bf{Proof}.}  
Recall that the updating equation of $\alpha_j,j=1,\cdots,p$ at each iteration are as follows:
\begin{equation} \label{alpha_jt}
	\alpha_j^{(t+1)} = \alpha_j^{(t)}  -  \eta \Big(\frac{1}{n}	\sum^{n}_{i=1}\nabla_{\alpha_j}
	L^{tr}_i(\alpha,\theta^{(t+1)})+ \lambda\frac{\alpha_j}{\|\alpha_j\|_2}\Big)\Big| _{\alpha_j^{(t)}}.
\end{equation}

Under the training dataset $\Phi_t$, \eqref{alpha_jt} can be rewritten as:
\begin{equation} \label{alpha_jt1}
	\alpha_j^{(t+1)} = \alpha_j^{(t)}  -   \eta_t  [\nabla_{\alpha_j}
	\mathcal{L}_n^{tr}(\alpha_j^{(t)},\theta^{(t+1)})|_{\Phi_t}
	+ \lambda\frac{\alpha_j^{(t)}}{\|\alpha_j^{(t)}\|_2}],
\end{equation}
where $\mathcal{L}_n^{tr}(\alpha^{(t)},\theta^{(t+1)})=\frac{1}{n}	\sum^{n}_{i=1}\nabla_{\alpha_j}
L^{tr}_i(\alpha,\theta^{(t+1)})$. 
In addition,  we can rewritten \eqref{alpha_jt1} as
\begin{equation*} 
	\alpha_j^{(t+1)} = \alpha_j^{(t)}  -   \eta_t[\nabla_{\alpha_j}
	\mathcal{L}_n^{tr}(\alpha_j^{(t)},\theta^{(t+1)}) + \varphi^{(t)} + \lambda\frac{\alpha_j^{(t)}}{\|\alpha_j^{(t)}\|_2}].
\end{equation*}
where $\varphi^{(t)} = \nabla_{\alpha_j} 
\mathcal{L}_n^{tr}(\alpha_j^{(t)},\theta^{(t+1)})| _{\Phi_t}-\nabla_{\alpha_j} 
\mathcal{L}_n^{tr}(\alpha_j^{(t)},\theta^{(t+1)})$. Since  $\Phi_t$ are drawn i.i.d. with finite  samples, $\varphi^{(t)}$  satisfies $ \mathbb{E}[\varphi^{(t)}]=0$ and  $\mathbb{E}[\|\varphi^{(t)}\|^2_2]\leq \sigma^2$. 

Observe that 
\begin{align} \label{Theorem2_1}
	&~~~\mathcal{L}^{tr}(\alpha_j^{(t+1)};\theta^{(t+1)}) - \mathcal{L}^{tr}({\alpha}_j^{(t)};\theta^{(t)})~\nonumber  \\
	&=\Big\{\mathcal{L}^{tr}( {\alpha}_j^{(t+1)};\theta^{(t+1)}) - \mathcal{L}^{tr}( {\alpha}_j^{(t+1)};\theta^{(t)}) \Big\}\\
 &~~~+\Big\{\mathcal{L}^{tr}( {\alpha}_j^{(t+1)};\theta^{(t)})- \mathcal{L}^{tr}( {\alpha}_j^{(t)};\theta^{(t)})\Big \}.\nonumber
\end{align} 

For the first term, based on the Lipschitz smoothness of   $\mathcal{L}^{tr}({\alpha}_j^{(t+1)};\theta^{(t+1)})$, we have 
	\begin{align*} 
		&~~~\mathcal{L}^{tr}({\alpha}_j^{(t+1)};\theta^{(t+1)}) - \mathcal{L}^{tr}(
		{\alpha}_j^{(t+1)};\theta^{(t)}) \nonumber \\
		&\leq   \langle \nabla_{\theta}\mathcal{L}_n^{tr}( {\alpha}_j^{(t+1)};\theta^{(t)}), \theta^{(t+1)}-\theta^{(t)}  \rangle + \frac{L}{2} \|\theta^{(t+1)} -\theta^{(t)}  \|^2_2 \nonumber \\	
		&=  \langle  \nabla_{\theta}\mathcal{L}_n^{tr}({\alpha}_j^{(t+1)};\theta^{(t)}),- \nu_t\nabla_{\theta}\mathcal{L}_m^{val}( \bar{\alpha}^{(t+1)},\theta^{(t)})  \rangle +\frac{L\nu_t^2}{2} \|\xi^{(t)}  \|^2_2 \nonumber\\
		&~~~- \nu_t \langle  \nabla_{\theta}\mathcal{L}_n^{tr}({\alpha}_j^{(t+1)};\theta^{(t)}), \xi^{(t)} \rangle + \frac{L\nu_t^2 }{2}  \| \nabla_{\theta} \mathcal{L}_m^{val}( \bar{\alpha}^{(t+1)};\theta^{(t)}) \| ^2_2 \nonumber\\
		&~~~ + L\nu_t^2 \langle  \nabla_{\theta}\mathcal{L}_m^{val}(\bar{\alpha}^{(t+1)};\theta^{(t)}), \xi^{(t)} \rangle.&	 \nonumber
	\end{align*}
For the second term, we have 

	\begin{align*} 
		& ~~~\mathcal{L}^{tr}({\alpha}_j^{(t+1)};\theta^{(t)}) - \mathcal{L}^{tr}(
		{\alpha}_j^{(t)};\theta^{(t)}) \nonumber \\
		&\leq \langle \nabla_{\alpha_j}\mathcal{L}^{tr}(
		{\alpha}_j^{(t)};\theta^{(t)}), {\alpha}_j^{(t+1)}-{\alpha}_j^{(t)}  \rangle + \frac{L}{2} \| {\alpha}_j^{(t+1)} - {\alpha}_j^{(t)}  \|^2_2    \\
		&=  (\frac{L\eta_t^2}{2}-\eta_t) \| \nabla_{\alpha_j}
		\mathcal{L}_n^{tr}(\alpha_j^{(t)},\theta^{(t)})  \|^2_2 + (L\eta_t^2-\eta_t) \langle \nabla_{\alpha_j}
		\mathcal{L}_n^{tr}(\alpha_j^{(t)},\theta^{(t)})	,\varphi^{(t)}  \rangle	\nonumber\\
		&~~~ + (L\eta_t^2-2\eta_t)  \langle \nabla_{\alpha_j}
		\mathcal{L}_n^{tr}(\alpha_j^{(t)},\theta^{(t)})	,\frac{\lambda\alpha_j^{(t)}}{\|\alpha_j^{(t)}\|_2}  \rangle + \frac{L\eta_t^2\lambda^2}{2}
  \nonumber\\
		&~~~ +(L\eta_t^2-2\eta_t)  \langle \varphi^{(t)}	,\frac{\lambda\alpha_j^{(t)}}{\|\alpha_j^{(t)}\|_2}  \rangle -\eta_t\lambda^2+ \frac{L\eta_t^2}{2}  \| \varphi^{(t)} \|^2_2 \nonumber\\
		&\leq   (L\eta_t^2-2\eta_t)  \| \nabla_{\alpha_j}
		\mathcal{L}_n^{tr}(\alpha_j^{(t)},\theta^{(t)})  \|^2_2 + (L\eta_t^2-\eta_t) \langle \nabla_{\alpha_j}
		\mathcal{L}_n^{tr}(\alpha_j^{(t)},\theta^{(t)})	,\varphi^{(t)}  \rangle	\nonumber\\		
		&+(L\eta_t^2-2\eta_t)  \langle \varphi^{(t)}	,\frac{\lambda\alpha_j^{(t)}}{\|\alpha_j^{(t)}\|_2}  \rangle + \frac{L\eta_t^2}{2} \| \varphi^{(t)}\|^2_2 + L\eta_t^2\lambda^2-2 \eta_t\lambda^2, &\nonumber
	\end{align*}

where the last inequality holds form the fact that $ \langle \nabla_{\alpha_j}
\mathcal{L}_n^{tr}(\alpha_j^{(t)},\theta^{(t)})	,\frac{\lambda\alpha_j^{(t)}}{\|\alpha_j^{(t)}\|_2}  \rangle \leq \frac{1}{2}( \|\nabla_{\alpha_j}
\mathcal{L}_n^{tr}(\alpha_j^{(t)},\theta^{(t)}) \|^2_2+\lambda^2) $.

Therefore, combining the above inequalites, we obtain

	\begin{align*}
		&~~~\mathcal{L}^{tr}(\alpha_j^{(t+1)};\theta^{(t+1)}) - \mathcal{L}^{tr}({\alpha}_j^{(t)};\theta^{(t)})~\nonumber  \\
		&\leq  \langle  \nabla_{\theta}\mathcal{L}_n^{tr}({\alpha}_j^{(t+1)};\theta^{(t)}),- \nu_t\nabla_{\theta}\mathcal{L}_m^{val}( \bar{\alpha}^{(t+1)},\theta^{(t)}) \rangle +\frac{L\nu_t^2}{2} \|\xi^{(t)} \|^2_2 \nonumber\\
		&~~~ - \nu_t \langle  \nabla_{\theta}\mathcal{L}_n^{tr}({\alpha}_j^{(t+1)};\theta^{(t)}), \xi^{(t)} \rangle + \frac{L\nu_t^2 }{2}  \| \nabla_{\theta} \mathcal{L}_m^{val}( \bar{\alpha}^{(t+1)};\theta^{(t)}) \| ^2_2 \nonumber\\
		&~~~ + L\nu_t^2 \langle  \nabla_{\theta}\mathcal{L}_m^{val}(\bar{\alpha}^{(t+1)};\theta^{(t)}), \xi^{(t)} \rangle  + (L\eta_t^2-2\eta_t)  \| \nabla_{\alpha_j}
		\mathcal{L}_n^{tr}(\alpha_j^{(t)},\theta^{(t)})  \|^2_2\nonumber\\	
		&~~~  + (L\eta_t^2-\eta_t) \langle \nabla_{\alpha_j}
		\mathcal{L}_n^{tr}(\alpha_j^{(t)},\theta^{(t)})	,\varphi^{(t)}  \rangle+ \frac{L\eta_t^2}{2} \| \varphi^{(t)}\|^2_2	\nonumber\\		
		&~~~+(L\eta_t^2-2\eta_t)  \langle \varphi^{(t)}	,\frac{\lambda\alpha_j^{(t)}}{\|\alpha_j^{(t)}\|_2}  \rangle  + L\eta_t^2\lambda^2-2 \eta_t\lambda^2, \nonumber
	\end{align*}

Since $\mathbb{E}[\xi^{(t)}]=0$, $\mathbb{E}[\|\xi^{(t)}\|^2_2]<\sigma^2$, $\mathbb{E}[\varphi^{(t)}]=0$ and  $\mathbb{E}[\|\varphi^{(t)}\|^2_2]<\sigma^2$ , taking expectation on both sides of the above inequality,  we can deduce that 
\begin{align} \label{expet_tr}
	&~~~\mathbb{E}[\mathcal{L}^{tr}(\alpha_j^{(t+1)};\theta^{(t+1)})] - \mathbb{E}[\mathcal{L}^{tr}({\alpha}_j^{(t)};\theta^{(t)})]\nonumber  \\
	&\leq \mathbb{E}[\langle  \nabla_{\theta}\mathcal{L}_n^{tr}({\alpha}_j^{(t+1)};\theta^{(t)}),- \nu_t\nabla_{\theta}\mathcal{L}_m^{val}( \bar{\alpha}^{(t+1)},\theta^{(t)}) \rangle] \\
	&~~~ +\frac{L\nu_t^2(\sigma^2+\rho ^2 )}{2} + (L\eta_t^2-2\eta_t)  \mathbb{E}[\| \nabla_{\alpha_j}
	\mathcal{L}_n^{tr}(\alpha_j^{(t)},\theta^{(t)})  \|^2_2]\nonumber\\	
	&~~~  +  \frac{L\eta_t^2(\sigma^2 +2\lambda^2)}{2} -2 \eta_t\lambda^2, \nonumber
\end{align}

Summing up \eqref{expet_tr} over $t=1,\cdots,T$ on both sides, we have 
	\begin{align*} 
		&~~~\sum^{T}_{t=1}(2\eta_t -  {L\eta_t^2} )\mathbb{E}[ \|\nabla_{\alpha_j} \mathcal{L}_n^{tr}({\alpha}_j^{(t)};\theta^{(t)}) \|^2_2]+ 2\lambda^2 \sum^{T}_{t=1}\eta_t \\
		&\leq \mathbb{E}[\mathcal{L}^{tr}(\alpha_j^{(1)};\theta^{(1)})]+  \frac{L(\sigma^2+\rho ^2 )}{2} \sum^{T}_{t=1}\nu_t^2 
		+	\frac{L\eta_t^2(\sigma^2 +2\lambda^2)}{2} \sum^{T}_{t=1}\eta_t^2	\\	
		&~~~ + \rho \sum^{T}_{t=1}\nu_t\|\nabla_{\theta}\mathcal{L}_m^{val}( \bar{\alpha}^{(t+1)},\theta^{(t)}) \|_2 < \infty,&
	\end{align*}
where the last inequality holds since $\sum^{T}_{t=1}\eta^2_t<\infty$, $\sum^{T}_{t=1}\nu^2_t<\infty$ and $\sum^{T}_{t=1}\nu_t\|\nabla_{\theta}\mathcal{L}_m^{val}( \bar{\alpha}^{(t+1)},\theta^{(t)}) \|_2<\infty$.

Hence, we have  

	\begin{align*} 	
		&~~~	\min\limits_{1\leq t\leq T} \mathbb{E}[\|\nabla_{\alpha_j} \mathcal{L}^{tr}({\alpha}_j^{(t+1)};\theta^{(t)})\|^2_2]\\
		&= \min\limits_{1\leq t\leq T} \mathbb{E}\Big[\Big\|\nabla_{\alpha_j} \mathcal{L}_n^{tr}({\alpha}_j^{(t+1)};\theta^{(t)})+\lambda\frac{\alpha_j^{(t)}}{\|\alpha_j^{(t)}\|_2}\Big\|^2_2\Big]\\	
		&\leq \frac{2\sum^{T}_{t=1}(2\eta_t -  {L\eta_t^2})\mathbb{E}[\|\nabla_{\alpha_j} \mathcal{L}_n^{tr}({\alpha}_j^{(t+1)};\theta^{(t)}) \|^2_2] +2\lambda^2\sum^{T}_{t=1}(2\eta_t - L\eta_t^2)} {\sum^{T}_{t=1}(2\eta_t - L\eta_t^2)}\\	
		&\leq \frac{2\sum^{T}_{t=1}(2\eta_t -  {L\eta_t^2})\mathbb{E}[\|\nabla_{\alpha_j} \mathcal{L}_n^{tr}({\alpha}_j^{(t+1)};\theta^{(t)}) \|^2_2] + 4\lambda^2\sum^{T}_{t=1}\eta_t}  {\sum^{T}_{t=1}\eta_t } \\
		&\leq \frac{2\sum^{T}_{t=1}(2\eta_t -  {L\eta_t^2})\mathbb{E}[\|\nabla_{\alpha_j} \mathcal{L}_n^{tr}({\alpha}_j^{(t+1)};\theta^{(t)}) \|^2_2] + 4\lambda^2\sum^{T}_{t=1}\eta_t}  {T } \\
         &~~~~\cdot  \max\{ L, \frac{\sqrt{T}}{C}\} = \mathcal{O}(\frac{C}{\sqrt{T}}).	&
	\end{align*}
  
 The second inequality holds since $\sum^{T}_{t=1}(2\eta_t - L\eta_t^2)=\sum^{T}_{t=1}\eta_t(2- L\eta_t)\geq\sum^{T}_{t=1}\eta_t$.  Hence, we can deduce that GSAMUL can always achieve $\min\limits_{1\leq t\leq T} \mathbb{E}[\|\nabla_{\alpha_j} \mathcal{L}^{tr}({\alpha}_j^{(t+1)};\theta^{(t)})\|^2_2]\leq\mathcal{O}(\frac{C}{\sqrt{T}})$ in $T$ steps. This completes the proof of Theorem \ref{Theorem2}. $\blacksquare $

\begin{table}[htbp]
	\footnotesize \centering \caption{Component functions and link functions in Example A and Example B}\label{simulation} \renewcommand \arraystretch{1.8}	
	\begin{tabular}{|p{1cm}|p{2.3cm}|p{4cm}|} \hline
		Functions &Example A   &Example B     \tabularnewline\hline		$f_j$		
		&$f_1(X_1)=\sin(\pi \cdot X_1$)   &$f_1(X_1)=0.3(\sin(X_1\cdot\pi)-\frac{2}{\pi})$\tabularnewline
		&$f_2(X_2)=0.5X_2^2-\frac{2}{3}$    &$f_2(X_2)=0.5((X_2-0.5)^2 -  \frac{1}{12})$\tabularnewline
		& &$f_3(X_3)=0.4(\exp(-X_3)+\exp(1)-1)$ \tabularnewline            
		& &$f_4(X_4)=\ln(2)-\frac{1}{1+X_4}$  \tabularnewline\hline
		$f$ &  $f(X)=\sum\limits_{j=1}^{2}f_j(X_j)$  & $f(X)=\sum\limits_{j=1}^{4}f_j(X_j)$ \tabularnewline\hline
		$g$    &$g(f)= 3\sin(f)$  &$g(f)=\exp(0.25f)$         \tabularnewline\hline
		Noise   &$\mathcal{N}(0,0.1)$  &$\mathcal{N}(0,0.1)$   \tabularnewline\hline
	\end{tabular}
\end{table}

\section{Experiments} \label{Experiments}

In this section, we evaluate the performance of the proposed GSAMUL on synthetic and real world datasets. We compare our method with the following methods: SpAM\cite{Ravikumar09}, GLMUL\cite{Muggeo2008Fitting}, GAMUL\cite{Lin2018Efficient}, PLAMUL\cite{Zhang2018Partially}, in which all methods are implemented by MATLAB. Moreover, according to the estimation method for GAMUL and GLMUL, they may not be suitable to combine the proposed method with a penalty to simultaneously estimate and select features. Hence, we only present the performance of estimating the link function when conducting the feature selection.

In experiment, a single hidden layer feedforward neural network is employed in the proposed model. We search the regularization parameter $\lambda$, the order of B-splines and the nodes of hidden layer in the range of $\{10^{-3}, 10^{-2},\cdots, 10^2, 10^3\}$, $\{3,4,\cdots, 10\}$ and  $\{5,7,\cdots, 50\}$, respectively. 

\begin{table*}[htbp]
	\scriptsize \centering \caption{Performance comparison on synthetic data (ASEs are in the scale of  $10^{-1}$ and $10^{-2}$ for Example A and B, respectively). }\label{dataset}\renewcommand \arraystretch{1.3}
	\begin{tabular}{p{1cm}p{1.1cm}p{1cm}p{1cm}p{1.5cm}|p{1cm}p{1cm}p{1cm}p{1cm}p{1.5cm}} \hline
		&	                    &\multicolumn{3}{c|}{Example A}	   &\multicolumn{5}{c}{Example B}  \tabularnewline\cline{3-4}\cline{5-10} 
		&Methods	&$f_1$	            &$f_2$					&g 			&$f_1$				&$f_2$				& $f_3$			&$f_4$   &g 	        \tabularnewline\hline	
		$n=300$&SpAM   &0.393 &0.460 &0.035(0.012) &0.710 &0.366 &0.377 &0.122 &0.043(0.011)\tabularnewline
		&GLMUL  &0.380 &0.271 &0.038(0.007) &0.910 &0.213 &0.906 &0.789 &0.032(0.016)\tabularnewline
		&GAMUL  &0.325 &0.346 &0.034(0.006) &0.841 &0.193 &0.882 &0.671 &0.029(0.008)\tabularnewline
		&PLAMUL &0.270 &0.420 &0.032(0.005) &0.560 &0.208 &0.752 &0.177 &0.026(0.015)\tabularnewline
		&GSAMUL &0.190 &0.206 &0.027(0.010) &0.726 &0.173 &0.890 &0.120 &0.023(0.013)\tabularnewline\hline
		$n=500$ &SpAM  &0.406 &0.470 &0.027(0.008) &0.532 &0.241 &0.418 &0.400 &0.027(0.007)\tabularnewline
		&GLMUL  &0.384 &0.275 &0.030(0.008) &0.888 &0.159 &0.463 &0.743 &0.032(0.019)\tabularnewline
		&GAMUL  &0.359 &0.361 &0.026(0.008) &0.520 &0.142 &0.448 &0.733 &0.025(0.011)\tabularnewline
		&PLAMUL &0.335 &0.447 &0.022(0.009) &0.216 &0.125 &0.434 &0.723 &0.022(0.014)\tabularnewline
		&GSAMUL &0.314 &0.189 &0.020(0.002) &0.605 &0.150 &0.457 &0.970 &0.017(0.008)\tabularnewline\hline
		$n=1000$&SpAM   &0.406 &0.477 &0.024(0.003)&0.917 &0.161 &0.377 &0.977&0.019(0.003)\tabularnewline
		&GLMUL  &0.384 &0.271 &0.028(0.004)&0.899 &0.141 &0.571 &0.723 &0.021(0.082)\tabularnewline
		&GAMUL  &0.356 &0.220 &0.024(0.005)&0.750 &0.122 &0.521 &0.852 &0.019(0.015)\tabularnewline
		&PLAMUL &0.329 &0.169 &0.020(0.006)&0.601 &0.104 &0.470 &0.981 &0.018(0.021)\tabularnewline
		&GSAMUL &0.349 &0.252 &0.018(0.003)&0.591 &0.104 &0.488 &0.928 &0.016(0.009)\tabularnewline\hline
	\end{tabular}
\end{table*} 

For the synthetic data, the average squared error (ASE) is used to evaluate the prediction results for $g(\cdot)$ and $f_j(\cdot)$, which is defined as 
\begin{align*}
	{ASE}_g &=  \frac{1}{n}\sum_{i=1}^{n}(g^*(f(x_i))-g_\mathbf{z}(f(x_i)))^2, \\
	ASE_j &=  \frac{1}{n}\sum_{i=1}^{n}(f_j^*(x_{ij})-f_{\mathbf{z},j}(x_{ij}))^2,&
\end{align*}
where $g_{\mathbf{z}}(\cdot)$ and $f_{\mathbf{z},j}(\cdot) $ are  estimated by GSAMUL,  and $g^*(\cdot)$ and $f^*_{j}(\cdot)$ are the real link function and component functions, respectively.

For the real-world data, we randomly select $40\%$ of the dataset as training set, $40\%$ as validating set and the remaining as testing set. Since the true useful features are unknown, we simply measure the prediction error via the relative sum of the squared error (RSSE), which is defined as 
\begin{equation*}
	RSSE =  \sum\limits_{x,y \in X_{tes}}(y - g_\mathbf{z}(f(x)))^2 / \sum\limits_{y \in X_{tes}}(y - \mathbb{E} y )^2,
\end{equation*}
where $g_\mathbf{z}(f(x))$ is the estimator of GSAMUL, $y$ is the output of $x$ in the testing dataset, and $\mathbb{E}y$ is the average of  $y$.

\subsection{Simulated data analysis}

\begin{table*}[htbp]
	\scriptsize \centering \caption{Results of variable selection for Example A and B. The ASE(Std) are in the scale of $10^{-2}$ and $10^{-3}$, respectively. }\label{resultSele}\renewcommand \arraystretch{1.3}
	\begin{tabular}{p{1cm}p{1.1cm}p{0.8cm}p{0.8cm}p{0.8cm}p{1.6cm}|p{0.8cm}p{0.8cm}p{0.8cm}p{1.6cm}} \hline
		&	     	   &\multicolumn{4}{c|}{Example A}	   &\multicolumn{4}{c}{Example B} \tabularnewline\cline{3-10} 
		&Methods	&Size &TP  &FP  &ASE(Std)	&Size	 &TP	&FP  &ASE(Std)	        \tabularnewline\hline	
		$p=20$   &SpAM	&2.00   &2.00	&0.00	&0.271(0.080)	&4.20	&3.85	&0.35	&0.332(0.117)\tabularnewline
		&GLMUL  &-&-&-&0.401(0.121) &-&-&-& 0.591(0.215) \tabularnewline
		&GAMUL &-&-&-&0.359(0.224) &-&-&-&0.473(0.325) \tabularnewline
		&PLAMUL	&2.00	&2.00	&0.00 	&0.252(0.122)	&4.15	&3.90	&0.25	&0.267(0.173)\tabularnewline
		&GSAMUL	&2.00	&2.00	&0.00	&0.206(0.029)	&4.20	&3.95  &0.25	&0.205(0.110) \tabularnewline\hline
		$p=40$  &SpAM	&2.00	&2.00	&0.00	&0.256(0.056)	&4.20	&3.75	&0.45	&0.346(0.127)\tabularnewline
		&GLMUL  &-&-&-&0.630(0.371) &-&-&-& 0.729(0.272) \tabularnewline
		&GAMUL &-&-&-&0.502(0.313)&-&-&-&0.646(0.314) \tabularnewline
		&PLAMUL	&2.00	&2.00	&0.00	&0.254(0.188)	&4.10	&3.75	&0.35	&0.273(0.202)\tabularnewline
		&GSAMUL	&2.00	&2.00	&0.00	&0.209(0.034)	&4.10	&3.90	&0.20	&0.232(0.095) 	\tabularnewline\hline
		$p=80$	&SpAM   &2.00	&2.00	&0.00	&0.258(0.066)	&4.20	&3.70	&0.50	&0.375(0.131)\tabularnewline
		&GLMUL  &-&-&-&0.771(0.438) &-&-&-&1.165(0.708)        
		\tabularnewline
		&GAMUL &-&-&-&0.611(0.324)&-&-&-&0.785(0.711)\tabularnewline
		&PLAMUL	&2.00	&2.00	&0.00	&0.236(0.103)  	&4.15	&3.75	&0.40	&0.253(0.117)\tabularnewline
		&GSAMUL &2.00	&2.00	&0.00	&0.210(0.029)	&4.15 	&3.80 & 0.35	&0.242(0.069)	\tabularnewline\hline								
	\end{tabular}
\end{table*}

Following the simulations in \cite{Zhang2018Partially} and \cite{Lin2018Efficient}, two examples (called Example A and Example B) are constructed. The training set is independently generated by the model 
\begin{equation*}
	Y = g(f_1(X_1) + f_2(X_2)+ \cdots + f_p(X_p)) + \epsilon, 
\end{equation*} 
where $\epsilon$ is the Gaussian noise. The validation set and the testing set are generated by $ g(\cdot) $ without noise. The input $X = (X_{1},\cdots, X_{p})^T$ is independently drawn from a uniform distribution $U(0, 1)$. Data details are presented in Table  \ref{simulation}. In Example A, the additive function $f(X)$ is formed by two component functions. In Example B, the additive function $f(X)$ is constructed by four component functions.


Firstly, we consider the scenarios in which there is no redundant variables (p = 2 for Example A, p = 4 for Example B) and the sample size $n = 300,~ 500,~1000$, respectively. For each scenario, we repeat 20 times to reduce the influence of random samples, and report the mean and standard deviation of ASE for $g_{\mathbf{z}}(\cdot) $ and $f_{\mathbf{z},j}(\cdot), j=1....,p$ in Table \ref{dataset}. Compared with SpAM, other methods except GLMUL have better performance in most cases. The result shows that the estimated link function is better than the given identify function. For the estimation of component functions in additive part, the results of all methods are similar, but GSAMUL has lowest prediction error than other methods for the estimation of link function.

\begin{figure}[htbp]
	\centering
	\includegraphics[width= 2.5 in]{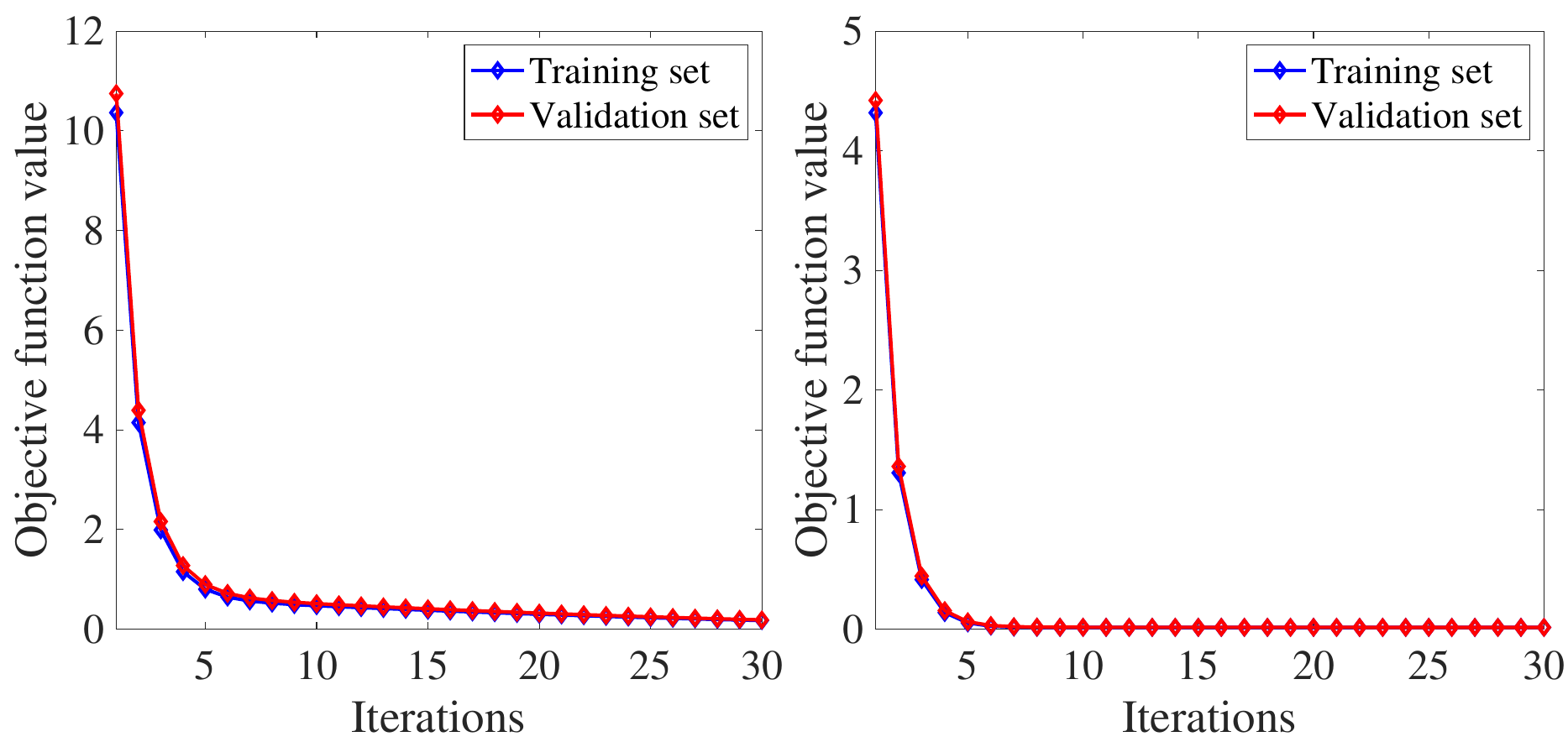}
	\caption{The convergence curves of GSAMUL for Example A (left) and Example B (right).}
	\label{Covergence}
\end{figure}

\begin{figure}[htbp]
	\centering
	\subfloat{\includegraphics[width= 3.3 in]{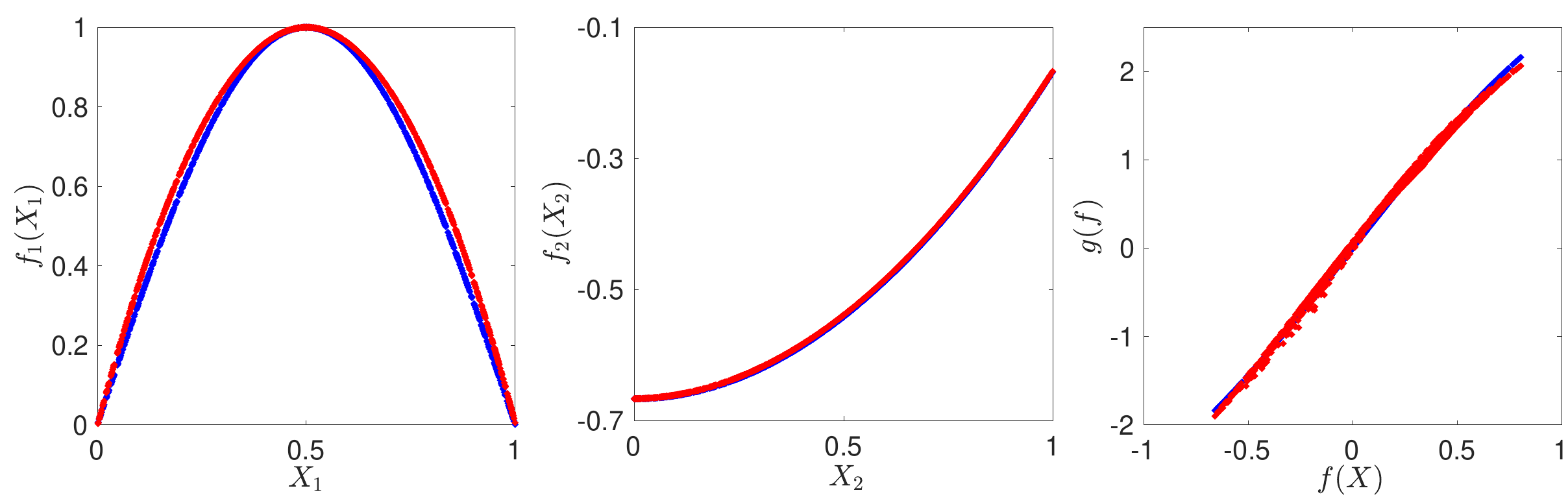}}
	\hfil \hspace{0pt} 
	\subfloat{\includegraphics[width= 3.3 in]{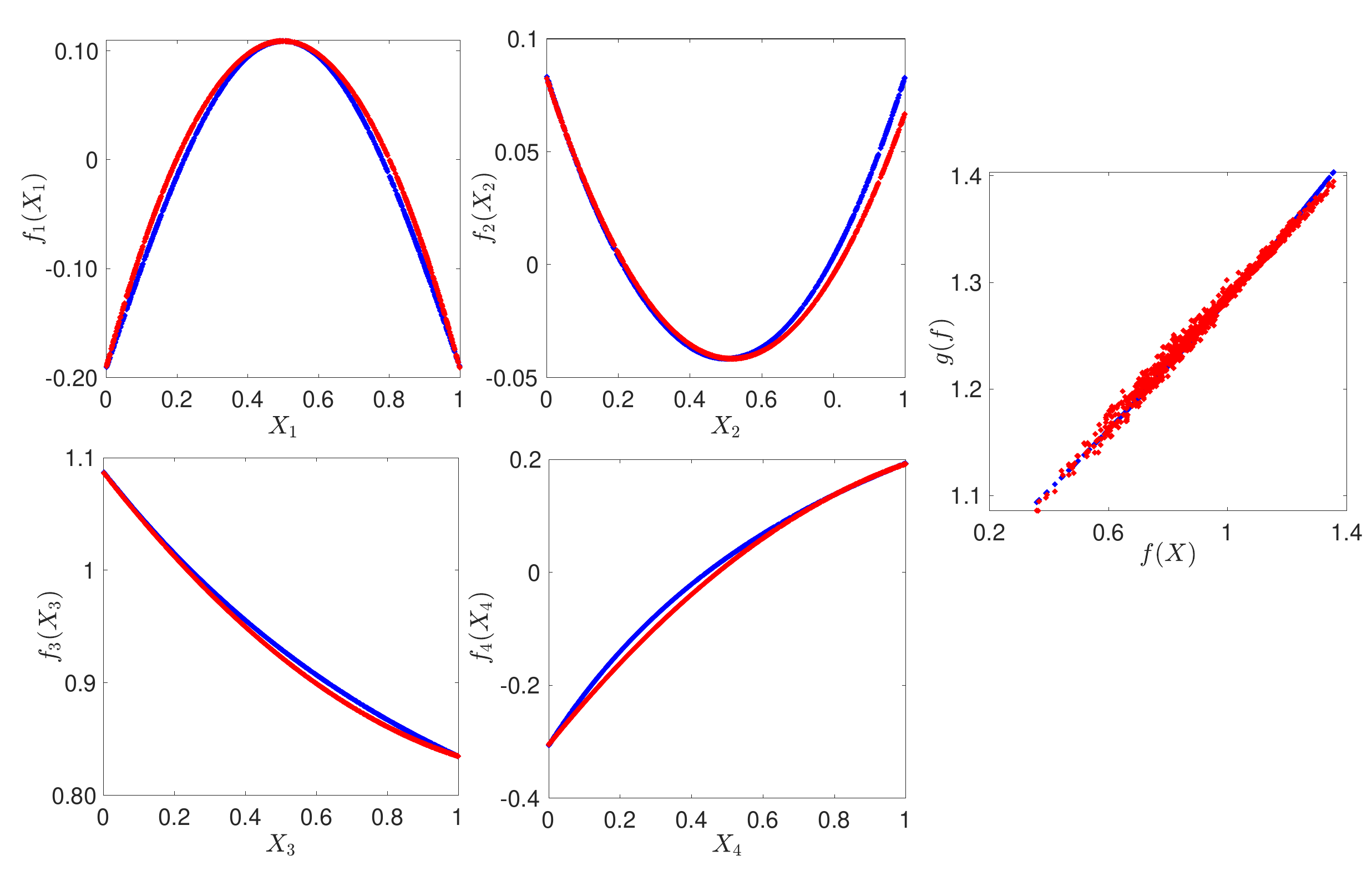}}	
	\caption{
		The estimates of component curves and link function (red- GSAMUL estimator; blue-true functions) for Example A (top) and Example B (bottom).}\label{estimationPlot}
\end{figure}

To have a visible idea about how well the proposed GSAMUL works, we present the plot for the estimated (red) and ture (blue) component and link functions in Fig. \ref{estimationPlot} . The result in  Fig. \ref{estimationPlot} shows that the estimated component and link functions by GSAMUL are very close to the true functions.

Furthermore, we present the effectiveness of GSAMUL on screening out the active variables in Table \ref{resultSele}. Since GLMUL and GAMUL do not consider the feature selection, their performance become worse with the increasing of the irrelative variables. For Example A with simple additive components, SpAM, PLAMUL and GSAMUL all can screen out the true variables correctly. And, for Example B with complicated additive components, GSAMUL outperforms other methods especially with more irrelative variables. Those empirical results show that  GSAMUL has lower ASE and better variable selection ability compared with SpAM and PLAMUL.


In addition, we conduct some experiments to verify the convergence of the optimization algorithm of GSAMUL. Fig. \ref{Covergence} shows the curve of objective function value of GSAMUL on the training set and validation set during the iterations. As can be seen, the iterative errors reduce quickly from the 1-th iteration to the 2-th iteration. And, the convergence can be achieved after about 30 iterations. As the number of iterations increases, the objective function values of Example A and  B on both the training set and the validation set gradually decrease, and finally reach a stationary convergence state.


\subsection{Real data analysis}

Now, we evaluate GSAMUL and related methods on Boston using census data, Plasma Retinol data, and Ozone data, which are widely used machine learning datasets \cite{Yang16, WangY21, Agarwal21}. Due to the limited features in real data, 20 irrelative variables are added to each data set, which are generated from the uniform distribution $U(-0.5,0.5)$. 

Boston using census data consists of 506 measurements with 13 variables, including CRIM, ZN, INDUS, CHAS, NOX, RM, AGE, DIS, RAD, TAX, PTRATIO, B and LSTAT. These variables are used to predict the median value of owner-occupied homesv (MEDV). Plasma Retinol data contains 315 observations with 12 variables, including AGE, SEX, SMOK, QUET, VIT, CAL, FAT, FIBER, ALCOHOL, CHOLES, BETA, and RET. It aims to establish the relationship between the plasma concentration of micronutrients and the related characteristics. Ozone data contains 366 observations with 11 input variables including  M, DM, DW, VDHT,  WDSP, HMDT, SBTP, IBTP, IBHT, DGPG and VSTY. Those variables are used to predict the upland ozone concentration (UPO3).


\begin{table*}[htbp]
	\scriptsize \centering \caption{Results of variable selection and RSSE (Std) on Boston housing, plasma retinol, and ozone data.} \label{SelectionReal}\renewcommand \arraystretch{1.3}
	\begin{tabular}{p{1.5cm}|p{0.7cm}<{\centering}p{0.4cm}<{\centering}p{0.8cm}<{\centering} p{0.7cm}<{\centering}p{0.7cm}<{\centering}p{0.4cm}<{\centering} p{0.5cm}<{\centering} p{0.3cm}<{\centering} p{0.5cm}<{\centering} p{0.5cm}<{\centering} p{1cm}<{\centering} p{0.3cm}<{\centering}p{0.5cm}<{\centering}p{1.5cm}<{\centering}} \hline
		&	\multicolumn{14}{c}{Boston Housing} \tabularnewline\hline
		Variables	&	CRIM& ZN& INDUS& CHAS& NOX& RM& AGE& DIS& RAD& TAX& PTRATIO& B &LSTAT& RSSE(Std)\tabularnewline\hline
		SpAM   &$\surd$&-&-&-&$\surd$&$\surd$&-&$\surd$&-&-&$\surd$&$\surd$&$\surd$&0.237(0.099)\tabularnewline
		PLAMUL &$\surd$&-&-&-&$\surd$&$\surd$&-&$\surd$&$\surd$&-&$\surd$&-&$\surd$&0.219(0.075)\tabularnewline
		GSAMUL  &-&-&-&-&$\surd$&$\surd$&-&$\surd$&-&-&-&-&$\surd$&0.201(0.086) \tabularnewline\hline
	\end{tabular}	
	\begin{tabular}{p{1.5cm}|p{0.5cm}<{\centering}p{0.5cm}<{\centering}p{0.6cm}<{\centering} p{0.6cm}<{\centering}p{0.4cm}<{\centering}p{0.4cm}<{\centering} p{0.4cm}<{\centering} <{\centering} p{0.8cm}<{\centering} p{1.2cm}<{\centering} p{1.2cm}<{\centering} p{0.6cm}<{\centering}p{0.5cm}<{\centering}p{1.5cm}<{\centering}} \hline
		&	\multicolumn{13}{c}{Plasma Retinol} \tabularnewline\hline
		Variables&AGE &SEX&SMOK&QUET&VIT&CAL&FAT&FIBER&ALCOHOL&CHOLES&BETA&RET&RSSE(Std)\tabularnewline\hline
		SpAM   &$\surd$&-&-&-&-&-&-&-&$\surd$&$\surd$&$\surd$&-&1.082(0.070)\tabularnewline
		PLAMUL &$\surd$&-&-&-&-&-&-&-&$\surd$&$\surd$&-&-&1.010(0.079)\tabularnewline
		GSAMUL  &$\surd$&-&$\surd$&-&-&-&-&-&$\surd$&-&-&-&0.975(0.093) \tabularnewline\hline
	\end{tabular}			
	\begin{tabular}{p{1.5cm}|p{0.4cm}<{\centering}p{0.55cm}<{\centering}p{0.8cm}<{\centering} p{0.8cm}<{\centering}p{0.8cm}<{\centering}p{0.8cm}<{\centering} p{0.8cm}<{\centering} p{0.8cm}<{\centering} p{0.8cm}<{\centering} p{0.8cm}<{\centering} p{0.8cm}<{\centering} p{1.5cm}<{\centering}} \hline
		&	\multicolumn{12}{c}{ Ozone} \tabularnewline\hline
		Variables	&M &DM &DW &VDHT &WDSP &HMDT &SBTP &IBHT &DGPG &IBTP &VSTY &RSSE(Std)\tabularnewline\hline
		SpAM   &$\surd$&-&-&$\surd$&-&$\surd$&$\surd$&$\surd$&$\surd$&$\surd$&$\surd$&0.310(0.056))\tabularnewline
		PLAMUL &$\surd$&-&-&$\surd$&-&-&$\surd$&-&$\surd$&$\surd$&-&0.318(0.058)\tabularnewline
		GSAMUL  &$\surd$&-&-&-&-&-&$\surd$&-&$\surd$&$\surd$&-&0.284(0.047) \tabularnewline\hline
	\end{tabular}	
\end{table*}	

We present results of variable selection, RSSE and its standard error in Table \ref{SelectionReal}.  For the Boston Housing data, the proposed GSAMUL only selects four variables but can obtain a smaller RSSE than SpAM and PLAMUL. For the Plasma Retinol data, AGE and ALCOHOL are selected by the there methods. However, SpAM achieves a bigger prediction error than our method. In addition, SMOK is ignored by SpAM and PLAMUL but chosen by GSAMUL. Based on the competitive performance of GSAMUL, it indicates that SMOK maybe a key factor for prediction. For the Ozone data, M, SBTP, DGPG and IBTP are the most important variables pronounced in \cite{Fasshauer12,Chen20}, which can be selected by the abovementioned methods. Moreover, VDHT is ignored by GSAMUL but chosen by SpAM and PLAMUL, which indicates that VDHT may not be a key factor for prediction.  According to the results, we find that the proposed GSAMUL has better performance  than the competitors.



In addition, to verify that GSAMUL still works on high-dimensional data, we repeat the abovementioned experiments by replacing 20 irrelative variables with 50, 200 and 500 irrelative variables and present the average results in Table \ref{SelectionReal500}. For brevity, we only present the results of selected variables and corresponding RSSE(Std). From Table \ref{SelectionReal500}, the results of these cases are roughly consistent, which illustrates GSAMUL still works even if there are many irrelative variables.

\begin{table*}[htbp]
	\scriptsize \centering \caption{The average results on GSAMUL under different size of irrelative variables for Boston housing, plasma retinol, and ozone data.} \label{SelectionReal500}\renewcommand \arraystretch{1.3}
	\begin{tabular}{p{1cm}|p{0.7cm}<{\centering}p{0.4cm}<{\centering}p{0.8cm}<{\centering} p{0.7cm}<{\centering}p{0.7cm}<{\centering}p{0.4cm}<{\centering} p{0.5cm}<{\centering} p{0.3cm}<{\centering} p{0.5cm}<{\centering} p{0.5cm}<{\centering} p{1cm}<{\centering} p{0.3cm}<{\centering}p{0.5cm}<{\centering}p{1.5cm}<{\centering}} \hline
		\multirow{2}{*}{\shortstack{Irrelative \\ variables}} &\multicolumn{14}{c}{ Boston Housing} \tabularnewline \cline{2-15}
		&	CRIM& ZN& INDUS& CHAS& NOX& RM& AGE& DIS& RAD& TAX& PTRATIO& B &LSTAT& RSSE(Std)\tabularnewline\hline
		50 &-&-&-&-&$\surd$&$\surd$&-&$\surd$&-&-&-&-&$\surd$&0.201(0.063)\tabularnewline
		200&-&-&-&-&$\surd$&$\surd$&-&-&-&-&-&-&$\surd$&0.203(0.070)\tabularnewline
		500&-&-&-&-&$\surd$&$\surd$&-&$\surd$&-&-&-&-&$\surd$&0.206(0.096) \tabularnewline\hline	   
	\end{tabular}	
	\begin{tabular}{p{1cm}|p{0.55cm}<{\centering}p{0.5cm}<{\centering}p{0.6cm}<{\centering} p{0.6cm}<{\centering}p{0.4cm}<{\centering}p{0.4cm}<{\centering} p{0.4cm}<{\centering} <{\centering} p{0.8cm}<{\centering} p{1.2cm}<{\centering} p{1.2cm}<{\centering} p{0.6cm}<{\centering}p{0.5cm}<{\centering}p{1.5cm}<{\centering}}  
		\multirow{2}{*}{\shortstack{Irrelative \\ variables}}	&	\multicolumn{13}{c}{Plasma Retinol} \tabularnewline\cline{2-14}
		&AGE &SEX&SMOK&QUET&VIT&CAL&FAT&FIBER&ALCOHOL&CHOLES&BETA&RET&RSSE(Std)\tabularnewline\hline
		50&$\surd$&-&$\surd$&-&-&-&-&-&$\surd$&-&-&-&0.979(0.061)\tabularnewline
		200&$\surd$&-&$\surd$&-&-&-&-&-&$\surd$&-&-&-&0.980(0.073)\tabularnewline
		500&$\surd$&-&-&-&-&-&-&-&$\surd$&-&-&-&0.987(0.034) \tabularnewline\hline
	\end{tabular}			
	\begin{tabular}{p{1cm}|p{0.4cm}<{\centering}p{0.55cm}<{\centering}p{0.8cm}<{\centering} p{0.8cm}<{\centering}p{0.8cm}<{\centering}p{0.8cm}<{\centering} p{0.8cm}<{\centering} p{0.8cm}<{\centering} p{0.8cm}<{\centering} p{0.8cm}<{\centering} p{0.8cm}<{\centering} p{1.5cm}<{\centering}}
		\multirow{2}{*}{\shortstack{Irrelative \\ variables}}&	\multicolumn{12}{c}{Ozone} \tabularnewline\cline{2-13}
		&M &DM &DW &VDHT &WDSP &HMDT &SBTP &IBHT &DGPG &IBTP &VSTY &RSSE(Std)\tabularnewline\hline  50&$\surd$&-&-&-&-&-&$\surd$&-&$\surd$&$\surd$&-&0.288(0.030))\tabularnewline
		200&$\surd$&-&-&-&-&-&$\surd$&-&$\surd$&$\surd$&-&0.299(0.066)\tabularnewline
		500&$\surd$&-&-&$\surd$&-&-&$\surd$&-&$\surd$&$\surd$&-&0.295(0.046) \tabularnewline\hline
	\end{tabular}	
\end{table*}



	\section{Conclusions} \label{conclusions}
	This paper formulated a new generalized sparse additive model with unknown link function, in which the additive part is estimated via the B-spline basis, and the unknown link function is estimated via a MLP network. We present a method to estimate those functions by means of a bilevel optimization problem where the data is split into training set and validation set. In theory, we provide guarantees about the convergence of the  approximate procedure. In applications, the proposed model has shown the advanced empirical performance on both simulated and real-world datasets.

	\bibliographystyle{IEEEtran}
	\bibliography{bib.bib}

\end{document}